\documentclass[11pt]{article}

\usepackage[preprint]{acl}

\usepackage{times}
\usepackage{latexsym}
\usepackage{subcaption}
\usepackage{booktabs,tabularx,makecell,array,amssymb}

\usepackage[T1]{fontenc}
\usepackage[utf8]{inputenc}

\usepackage{microtype}

\usepackage{inconsolata}

\usepackage{graphicx}

\usepackage{algorithm}
\usepackage{algorithmic}
\usepackage{amsmath}

\usepackage{cleveref}
\usepackage{todonotes}

\title{BlackboxNLP-2025 MIB Shared Task: Exploring Ensemble Strategies for Circuit Localization Methods}

\author{
  \textbf{Philipp Mondorf},
  \textbf{Mingyang Wang},
  \textbf{Sebastian Gerstner},
  \textbf{Ahmad Dawar Hakimi},\\
  \textbf{Yihong Liu},
  \textbf{Leonor Veloso},
  \textbf{Shijia Zhou},\\
  \textbf{Hinrich Schütze},
  \textbf{Barbara Plank}
  \\
  LMU Munich \& Munich Center for Machine Learning
\\
  \small{
    \textbf{Correspondence:} \href{mailto:p.mondorf@lmu.de}{p.mondorf@lmu.de}
  }
}

\begin{document}
\maketitle
\begin{abstract}

The Circuit Localization track of the Mechanistic Interpretability Benchmark (MIB) evaluates methods for localizing \textit{circuits} within large language models (LLMs), i.e., subnetworks responsible for specific task behaviors. 
In this work, we investigate whether ensembling two or more circuit localization methods can improve performance.
We explore two variants: parallel and sequential ensembling.
In parallel ensembling, we combine attribution scores assigned to each edge by different methods—e.g., by averaging or taking the minimum or maximum value. In the sequential ensemble, we use edge attribution scores obtained via EAP-IG as a warm start for a more expensive but more precise circuit identification method, namely edge pruning. We observe that both approaches yield notable gains on the benchmark metrics, leading to a more precise circuit identification approach. Finally, we find that taking a parallel ensemble over various methods, including the sequential ensemble, achieves the best results. We evaluate our approach in the BlackboxNLP 2025 MIB Shared Task, comparing ensemble scores to official baselines across multiple model–task combinations.

\end{abstract}

\section{Introduction}
The Circuit Localization track of MIB \citep{mueller2025mib} evaluates the ability of interpretability methods to identify \emph{circuits}, i.e., subnetworks in language models that are responsible for solving a specific task.
Its performance metrics measure how well the subnetwork can perform the given task on its own.
As proposed by the shared task, this is measured in two specific ways: \textit{circuit performance ratio} (CPR) and \textit{circuit-model difference} (CMD). The first metric, CPR, measures how well the method finds performant circuits at various circuit sizes.
The second metric, CMD, measures how closely the circuit's behavior resembles the model's task-specific behavior at various circuit sizes \citep{mueller2025mib}. Typically, a circuit localization method gives an attribution score to each edge in the network; it is then possible to choose different thresholds for including an edge in the circuit. Therefore, CPR and CMD scores are integrated over different circuit sizes to give a more holistic picture of each method.

In this work, we investigate whether ensembling two or more baseline methods can lead to better performance.
We hypothesize that different methods might be biased towards different model components, and that ensembling them might yield a more robust picture. Specifically, we explore two variants of ensembling: parallel and sequential. In parallel ensembling, we gather the scores assigned to each edge by different methods, typically by averaging, but also by taking the maximum or minimum score. In sequential ensembling, we first warm-start with EAP-IG \citep{sundararajan2017ig,hanna2024have} attribution scores, then apply edge pruning with attribution score sign recovery.

We evaluate our methods on both the private and public test sets. Our sequential ensemble method with warm-start edge pruning (\texttt{s-ens}) improves performance across tasks over the EAP-IG-inputs baseline, while the parallel ensemble (\texttt{p-ens}) also yields consistent gains. Motivated by these results, we combine both strategies—integrating warm-start edge pruning with the three EAP-based methods for parallel ensembling—into \texttt{hybrid-ens}, a hybrid ensemble, which achieves the best overall performance on the leaderboard, with the lowest CMD score (lower is better) and highest CPR score (higher is better). Our code is publicly available.\footnote{https://github.com/cisnlp/MIB-circuit-track}

\section{Background}
In this section, we describe the models, tasks, and base methods used in our experiments for the shared task. 

\subsection{Models and tasks}\label{ss:models-tasks}
To showcase our method, we use the three mandatory model/task combinations: (1) GPT-2-Small \cite{radford2019language} on IOI (indirect object identification, \citealp{wang2023interpretability}); (2) Qwen-2.5-0.5B \cite{qwen2} on IOI; (3) Qwen-2.5-0.5B on MCQA (multiple choice question answering without task-specific knowledge, \citealp{wiegreffe2024answer}).

We use both datasets in the version provided by~\citet{mueller2025mib}. Each one has train/validation/test splits, plus a private test set that is used for the leaderboard.\footnote{\url{https://huggingface.co/spaces/mib-bench/leaderboard}}

\subsection{Base methods}
We experiment with various base methods in our parallel and sequential ensemble settings.

A commonly used method is \textit{edge activation patching} \cite{Vig2020,finlayson-etal-2021-causal}. It computes the effect of replacing the activation of an edge with its activation on the counterfactual input. This is very inefficient (it needs a separate forward pass for each edge), so various approximations have been proposed. We use the following three as base methods (as implemented by \cite{mueller2025mib}):
(1) EAP (\textit{edge attribution patching}, \citealp{nanda2023attribution,syed-etal-2024-attribution}), (2) EAP-IG-inputs (\textit{edge attribution patching with integrated gradients over inputs}, \citealp{sundararajan2017ig,hanna2024have}), and (3) EAP-IG-activations \cite{sundararajan2017ig,marks2024sparsefeaturecircuitsdiscovering}.
They are among the best performing methods according to \citet{mueller2025mib}, with EAP-IG-inputs being the best one overall, so we use EAP-IG-inputs as the baseline for comparison in \Cref{s:results}.

We further incorporate edge pruning \cite{NEURIPS2024_20fdaf67}, a more precise but computationally expensive method, into the ensemble framework.

\section{Method Overview}
We explore two variants of ensembling: parallel and sequential. In parallel ensembling, we gather the scores different methods give to any edge -- for example using an average over methods, but we also experiment with taking the maximum or minimum score. In sequential ensembling, we first use EAP-IG-inputs as a warm start, and then perform edge pruning.

\subsection{Parallel ensembling}
As implemented by \citet{mueller2025mib}, a circuit detection method (\textit{base method}) gives a score to each edge of the model. We pre-compute these scores for a range of methods. As a result, each edge gets different \textit{base scores}. Our goal is to consolidate these to one \textit{final score} for each edge.

\subsubsection{Comparing base scores across methods}
We first have to answer the question whether the base scores are comparable across methods.

The EAP variants (first three methods above) are all designed to be approximations of edge activation patching.
We therefore assume the scores to be comparable across these methods.

The original edge pruning method obtains a mask (i.e., importance score) for each edge in the range [0, 1] without the sign information. Therefore, the importance score value is not comparable with the EAP-based methods. To make the scores of different methods in a matching space, we use the method described in Section~\ref{ss:sequential} to recover the sign information for a better hybrid ensemble (detailed description in Section~\ref{ss:final-submit}).

\subsubsection{Reduction methods}
We experiment with four score reduction methods across base approaches: (1) mean, (2) weighted average, (3) maximum, and (4) minimum. Interpreting scores as binary indicators, the maximum corresponds to a union of circuits (edges selected by any method), while the minimum corresponds to an intersection (edges selected by all methods). Ultimately, taking the mean yields the best results.

\subsection{Sequential ensembling}\label{ss:sequential}
The main idea of our sequential ensemble approach is to use the attribution scores produced by a fast circuit identification method to warm-start a slower, more precise method, thereby achieving faster convergence and further refining the initial scores. Specifically, we use edge attribution values computed via EAP-IG-inputs~\citep{hanna2024have} to initialize the learnable mask\textemdash{}specifically, the learnable log alpha parameters\textemdash{}of edge pruning~\citep{NEURIPS2024_20fdaf67}.

We convert attribution scores to hard-concrete log alpha values as follows. First, we take the absolute values of the edge attribution scores. These unsigned importance scores are grouped per layer and rank-normalized within each group to $[0,1]$. We then fit a logistic mapping to these ranks so that its average ``keep'' probability matches a predefined starting edge sparsity (a hyperparameter), yielding per-edge keep probabilities. Finally, we invert the hard-concrete expectation to obtain log alpha values via the stretched-sigmoid inverse, with standard clipping.

Once we initialize the log alphas as described, we refine the scores via edge pruning~\citep{NEURIPS2024_20fdaf67}. After training, we recover signed attribution scores from the refined log alpha values. To this end, we first compute the mask $\mathbf{z}$ as defined by~\citet{NEURIPS2024_20fdaf67}:

\[
\begin{aligned}
\mathbf{u} &\sim \mathrm{Uniform}(\varepsilon,\, 1-\varepsilon)\\
\mathbf{s} &= \sigma\!\left(\frac{1}{\beta}\cdot\log \frac{\mathbf{u}}{1-\mathbf{u}} + \log \mathbf{\alpha}\right)\\
\tilde{\mathbf{s}} &= \mathbf{s} \times (r - l) + l\\
\mathbf{z} &= \min\!\bigl(1,\, \max(0,\, \tilde{\mathbf{s}})\bigr)
\end{aligned}
\]

where $\varepsilon = 10^{-6}$, $\frac{1}{\beta} = \frac{2}{3}$, and $[l, r] = [-0.1, 1.1]$. Note that the mask values are positive, i.e., $\mathbf{z} \in [0, 1]$, and thus contain no sign information (i.e., whether an edge contributes positively or negatively to the model’s behavior). To recover sign information, we experiment with two approaches:
\begin{enumerate}
\item Reuse the sign from the initial EAP-IG-inputs attribution values, or
\item Compute a forward pass of the model in which each edge's output is a weighted combination of its original activation and a reference activation, determined by the learned mask values

\[
\begin{aligned}
y_i &= f_i\Bigl(
  z_{0i}\,y_0 + (1 - z_{0i})\,\tilde{y}_0 \\
&\qquad + \sum_{\substack{1 \le j < i \\ j \text{ upstream of } i}}
      \bigl( z_{ji}\,y_j + (1 - z_{ji})\,\tilde{y}_j \bigr)
\Bigr).
\end{aligned}
\]

and then compute the gradients of a metric $M$ (e.g., the KL divergence between circuit and model) with respect to the mask $\frac{\partial M}{\partial z}$. The sign of this gradient indicates whether the edge is useful for the task behavior (positive) or not useful (negative). We call this approach \emph{z-score attribution}.
\end{enumerate}

When performing hyperparameter optimization, we include a boolean that specifies whether the first or second approach is used. More details on our hyperparameters and the optimization procedure can be found in Section~\ref{s:experimental-setup} and Appendix~\ref{app:hyperparameters}.

\begin{table*}[h!]
\footnotesize
    \caption{\textbf{CMD} scores (lower is better) on the \textbf{private} test set, i.e., on the final leaderboard.}
    \centering
    \begin{tabular}{c|cccc}
        \hline
        Method & GPT-IOI & Qwen-IOI & Qwen-MCQA & \textbf{Average} \\
        \hline
        EAP-IG-inputs & 0.02 & 0.01 & 0.05 & \textbf{0.03} \\
        \hline
        \texttt{s-ens} & 0.03 & 0.02 & 0.07 & 0.04 \\
        \texttt{p-ens} & 0.02 & 0.02 & 0.07 & 0.04 \\
        \texttt{hybrid-ens} & 0.03 & 0.02 & 0.04 & \textbf{0.03}\\    
        \hline
    \end{tabular}
    \label{tab:cmd-private}
\end{table*}

\begin{table*}[h!]
\footnotesize
    \caption{\textbf{CPR} scores (higher is better) on the \textbf{private} test set, i.e., on the final leaderboard.}
    \centering
    \begin{tabular}{c|cccc}
        \hline
        Method & GPT-IOI & Qwen-IOI & Qwen-MCQA & \textbf{Average} \\
        \hline
        EAP-IG-inputs & 1.89 & 1.73 & 1.15 & 1.59 \\
        \hline
        \texttt{s-ens} & 2.37 & 1.71 & 1.16 & 1.75 \\
        \texttt{p-ens} & 2.11 & 1.88 & 0.79 & 1.59 \\
        \texttt{hybrid-ens} & 2.43 & 1.88 & 1.19  & \textbf{1.83} \\
        \hline
    \end{tabular}
    \label{tab:cpr-private}
\end{table*}

\begin{table*}[h!]
\footnotesize
    \caption{\textbf{CMD} scores (lower is better) on the \textbf{public} test set. EAP-IG-inputs scores are copied from \citet{mueller2025mib}. Our own scores should be taken with a grain of salt because we could not precisely reproduce the baselines.}
    \centering
    \begin{tabular}{c|cccc}
        \hline
        Method & GPT-IOI & Qwen-IOI & Qwen-MCQA & \textbf{Average} \\
        \hline
        EAP-IG-inputs & 0.03 & 0.02 & 0.08 & 0.04 \\
        \hline
        \texttt{s-ens} & 0.03 & 0.01 & 0.09 & 0.04 \\
        \texttt{p-ens} & 0.03 & 0.03 & 0.09 & 0.05 \\
        \texttt{hybrid-ens} & 0.02 & 0.03 & 0.04 & \textbf{0.03} \\
        \hline
    \end{tabular}
    \label{tab:cmd-public}
\end{table*}

\begin{table*}[h!]
\footnotesize
    \caption{\textbf{CPR} scores (higher is better) on the \textbf{public} test set. EAP-IG-inputs scores are copied from \citet{mueller2025mib}. Our own scores should be taken with a grain of salt because we could not precisely reproduce the baselines.}
    \centering
    \begin{tabular}{c|cccc}
        \hline
        Method & GPT-IOI & Qwen-IOI & Qwen-MCQA & \textbf{Average} \\
        \hline
        EAP-IG-inputs & 1.85 & 1.63 & 1.16 & 1.55 \\
        \hline
        \texttt{s-ens} & 2.22 & 1.62 & 0.99 & 1.61 \\
        \texttt{p-ens} & 2.06 & 1.83 & 0.99 & 1.63 \\
        \texttt{hybrid-ens} & 2.23 & 1.83 & 1.22 & \textbf{1.76} \\
        \hline
    \end{tabular}
    \label{tab:cpr-public}
\end{table*}
\subsection{Final submissions}\label{ss:final-submit}
After preliminary experiments, we finally submitted three different versions of ensembling:

\textbf{\texttt{s-ens}} (sequential ensembling, i.e., warmstart edge pruning with Z-score attribution): We use edge attribution values computed via EAP-IG-inputs to initialize the learnable mask of edge pruning as described above in \cref{ss:sequential}.

\textbf{\texttt{p-ens}} (parallel ensembling):
We calculate the average scores (non-weighted) from the three EAP variants for all three model-task combinations.

\textbf{\texttt{hybrid-ens}}: We combine the parallel and sequential strategies by taking the unweighted average over all four methods—the three EAP variants and warm-start edge pruning—for all model–task combinations.

\section{Experimental Setup}\label{s:experimental-setup}
\paragraph{Warmstart edge pruning} Once we have initialized the log alpha from the EAP-IG-inputs attribution scores, as defined in Section~\ref{ss:sequential}, we train each model on its respective task via edge pruning~\citep{NEURIPS2024_20fdaf67} for 1,000 training steps, using a batch size of 20 for GPT-2 and 10 for Qwen-2.5. We search for hyperparameters via Bayesian optimization on the validation set, selecting the combination that minimizes the combined objective $P = \text{CMD} - \text{CPR}$. For more details on the final hyperparameters and the optimization procedure, see Appendix~\ref{app:hyperparameters}.

\section{Results}\label{s:results}
We evaluated our methods on both the private and public test sets of the shared task using the CMD (lower is better) and CPR (higher is better) metrics. \Cref{tab:cmd-private,tab:cpr-private,tab:cmd-public,tab:cpr-public} summarize the results.

On both test sets, we observe that our sequential ensemble method with warm-start edge pruning (\texttt{s-ens}) already improves the performance across tasks compared to the EAP-IG-inputs baseline, particularly with respect to the CPR metric, achieving a +0.16 average gain. The parallel ensemble (\texttt{p-ens}) also provides consistent improvements over the baseline.

Building on these findings, we combined both strategies—integrating warm-start edge pruning with the three EAP-based methods for parallel ensembling—into our final submission, \texttt{hybrid-ens}. This combined approach achieves the best overall results on the private set, with the lowest CMD score (0.03 average) and the highest CPR score (1.83 average) on the leaderboard.

\section{Conclusion}
In this work, we introduce our system for the MIB shared task, exploring ensemble strategies for combining various circuit localization methods. We investigate two complementary ensemble strategies: a sequential ensemble with warm-start edge pruning and a parallel ensemble over multiple EAP-based variants, each yielding improvements over the baseline. Combining these strategies into a hybrid ensemble achieves the lowest CMD scores and highest CPR scores on both private and public leaderboards, demonstrating that ensembling complementary circuit localization methods can consistently and robustly enhance performance in mechanistic interpretability benchmarks.

For future work, we plan to investigate more diverse base methods, develop adaptive weighting schemes for ensembling, design improved strategies to integrate edge pruning with attribution-based approaches, and optimize for efficiency to scale to larger models and more complex tasks.

\section*{Limitations}
While our hybrid ensemble achieves strong results on the MIB shared task, several limitations remain. Our current ensemble relies on a naïve equal-weight averaging of the base methods, which may not fully exploit their complementary strengths; more adaptive weighting strategies could potentially yield further improvements. Additionally, our experiments are restricted to the task–model combinations defined in the shared task, and it is unclear how well the ensemble strategies generalize to other interpretability benchmarks or model architectures. Finally, the integration of edge pruning with attribution-based approaches is still heuristic, leaving room for more principled formulations that could improve both interpretability and performance.

\section*{Acknowledgments}
We would like to thank Lea Hirlimann and Robert Litschko for their input and feedback throughout the project. We further acknowledge the support for BP through the ERC Consolidator Grant DIALECT 101043235 and for HS through the Deutsche Forschungsgemeinschaft (project SCHU 2246/14-1).

% Custom bibliography entries only
\bibliography{custom}

\begin{thebibliography}{14}
\providecommand{\natexlab}[1]{#1}

\bibitem[{Bhaskar et~al.(2024)Bhaskar, Wettig, Friedman, and Chen}]{NEURIPS2024_20fdaf67}
Adithya Bhaskar, Alexander Wettig, Dan Friedman, and Danqi Chen. 2024.
\newblock \href {https://proceedings.neurips.cc/paper_files/paper/2024/file/20fdaf67581e6d7157376d1ed584040a-Paper-Conference.pdf} {Finding transformer circuits with edge pruning}.
\newblock In \emph{Advances in Neural Information Processing Systems}, volume~37, pages 18506--18534. Curran Associates, Inc.

\bibitem[{Ferrando and Voita(2024)}]{Ferrando2024a}
Javier Ferrando and Elena Voita. 2024.
\newblock \href {https://doi.org/10.18653/v1/2024.emnlp-main.965} {Information flow routes: Automatically interpreting language models at scale}.
\newblock pages 17432--17445, Miami, Florida, USA.

\bibitem[{Finlayson et~al.(2021)Finlayson, Mueller, Gehrmann, Shieber, Linzen, and Belinkov}]{finlayson-etal-2021-causal}
Matthew Finlayson, Aaron Mueller, Sebastian Gehrmann, Stuart Shieber, Tal Linzen, and Yonatan Belinkov. 2021.
\newblock \href {https://doi.org/10.18653/v1/2021.acl-long.144} {Causal analysis of syntactic agreement mechanisms in neural language models}.
\newblock In \emph{Proceedings of the 59th Annual Meeting of the Association for Computational Linguistics and the 11th International Joint Conference on Natural Language Processing (Volume 1: Long Papers)}, pages 1828--1843, Online. Association for Computational Linguistics.

\bibitem[{Hanna et~al.(2024)Hanna, Pezzelle, and Belinkov}]{hanna2024have}
Michael Hanna, Sandro Pezzelle, and Yonatan Belinkov. 2024.
\newblock \href {https://openreview.net/forum?id=grXgesr5dT} {Have faith in faithfulness: Going beyond circuit overlap when finding model mechanisms}.
\newblock In \emph{ICML 2024 Workshop on Mechanistic Interpretability}.

\bibitem[{Marks et~al.(2025)Marks, Rager, Michaud, Belinkov, Bau, and Mueller}]{marks2024sparsefeaturecircuitsdiscovering}
Samuel Marks, Can Rager, Eric~J Michaud, Yonatan Belinkov, David Bau, and Aaron Mueller. 2025.
\newblock \href {https://openreview.net/forum?id=I4e82CIDxv} {Sparse feature circuits: Discovering and editing interpretable causal graphs in language models}.
\newblock In \emph{The Thirteenth International Conference on Learning Representations}.

\bibitem[{Mueller et~al.(2025)Mueller, Geiger, Wiegreffe, Arad, Arcuschin, Belfki, Chan, Fiotto-Kaufman, Haklay, Hanna, Huang, Gupta, Nikankin, Orgad, Prakash, Reusch, Sankaranarayanan, Shao, Stolfo, Tutek, Zur, Bau, and Belinkov}]{mueller2025mib}
Aaron Mueller, Atticus Geiger, Sarah Wiegreffe, Dana Arad, Iván Arcuschin, Adam Belfki, Yik~Siu Chan, Jaden Fiotto-Kaufman, Tal Haklay, Michael Hanna, Jing Huang, Rohan Gupta, Yaniv Nikankin, Hadas Orgad, Nikhil Prakash, Anja Reusch, Aruna Sankaranarayanan, Shun Shao, Alessandro Stolfo, and 4 others. 2025.
\newblock \href {https://arxiv.org/abs/2504.13151} {{MIB}: A mechanistic interpretability benchmark}.
\newblock \emph{Preprint}, arXiv:2504.13151.

\bibitem[{Nanda(2023)}]{nanda2023attribution}
Neel Nanda. 2023.
\newblock \href {https://www.neelnanda.io/mechanistic-interpretability/attribution-patching} {Attribution {Patching}: {Activation} {Patching} {At} {Industrial} {Scale}}.

\bibitem[{Radford et~al.(2019)Radford, Wu, Child, Luan, Amodei, Sutskever et~al.}]{radford2019language}
Alec Radford, Jeffrey Wu, Rewon Child, David Luan, Dario Amodei, Ilya Sutskever, and 1 others. 2019.
\newblock Language models are unsupervised multitask learners.
\newblock \emph{OpenAI blog}, 1(8):9.

\bibitem[{Sundararajan et~al.(2017)Sundararajan, Taly, and Yan}]{sundararajan2017ig}
Mukund Sundararajan, Ankur Taly, and Qiqi Yan. 2017.
\newblock Axiomatic attribution for deep networks.
\newblock In \emph{Proceedings of the 34th International Conference on Machine Learning - Volume 70}, ICML'17, page 3319–3328. JMLR.org.

\bibitem[{Syed et~al.(2024)Syed, Rager, and Conmy}]{syed-etal-2024-attribution}
Aaquib Syed, Can Rager, and Arthur Conmy. 2024.
\newblock \href {https://doi.org/10.18653/v1/2024.blackboxnlp-1.25} {Attribution patching outperforms automated circuit discovery}.
\newblock In \emph{Proceedings of the 7th BlackboxNLP Workshop: Analyzing and Interpreting Neural Networks for NLP}, pages 407--416, Miami, Florida, US. Association for Computational Linguistics.

\bibitem[{Vig et~al.(2020)Vig, Gehrmann, Belinkov, Qian, Nevo, Sakenis, Huang, Singer, and Shieber}]{Vig2020}
Jesse Vig, Sebastian Gehrmann, Yonatan Belinkov, Sharon Qian, Daniel Nevo, Simas Sakenis, Jason Huang, Yaron Singer, and Stuart Shieber. 2020.
\newblock Causal mediation analysis for interpreting neural nlp: The case of gender bias.
\newblock \emph{arXiv preprint arXiv:2004.12265}.

\bibitem[{Wang et~al.(2023)Wang, Variengien, Conmy, Shlegeris, and Steinhardt}]{wang2023interpretability}
Kevin~Ro Wang, Alexandre Variengien, Arthur Conmy, Buck Shlegeris, and Jacob Steinhardt. 2023.
\newblock \href {https://openreview.net/forum?id=NpsVSN6o4ul} {Interpretability in the wild: a circuit for indirect object identification in {GPT}-2 small}.
\newblock In \emph{The Eleventh International Conference on Learning Representations}.

\bibitem[{Wiegreffe et~al.(2025)Wiegreffe, Tafjord, Belinkov, Hajishirzi, and Sabharwal}]{wiegreffe2024answer}
Sarah Wiegreffe, Oyvind Tafjord, Yonatan Belinkov, Hannaneh Hajishirzi, and Ashish Sabharwal. 2025.
\newblock \href {https://openreview.net/forum?id=6NNA0MxhCH} {Answer, assemble, ace: Understanding how {LM}s answer multiple choice questions}.
\newblock In \emph{The Thirteenth International Conference on Learning Representations}.

\bibitem[{Yang et~al.(2024)Yang, Yang, Hui, Zheng, Yu, Zhou, Li, Li, Liu, Huang, Dong, Wei, Lin, Tang, Wang, Yang, Tu, Zhang, Ma, Xu, Zhou, Bai, He, Lin, Dang, Lu, Chen, Yang, Li, Xue, Ni, Zhang, Wang, Peng, Men, Gao, Lin, Wang, Bai, Tan, Zhu, Li, Liu, Ge, Deng, Zhou, Ren, Zhang, Wei, Ren, Fan, Yao, Zhang, Wan, Chu, Liu, Cui, Zhang, and Fan}]{qwen2}
An~Yang, Baosong Yang, Binyuan Hui, Bo~Zheng, Bowen Yu, Chang Zhou, Chengpeng Li, Chengyuan Li, Dayiheng Liu, Fei Huang, Guanting Dong, Haoran Wei, Huan Lin, Jialong Tang, Jialin Wang, Jian Yang, Jianhong Tu, Jianwei Zhang, Jianxin Ma, and 40 others. 2024.
\newblock \href {https://arxiv.org/pdf/2407.10671} {Qwen2 technical report}.
\newblock \emph{arXiv preprint arXiv:2407.10671}.

\end{thebibliography}

\appendix

\section{Hyperparameters}
\label{app:hyperparameters}
In Table~\ref{tab:final-hparams}, we report the final hyperparameters for each model and task, as described in Section Section~\ref{ss:models-tasks}. These were selected based on a hyperparameter search via Bayesian optimization over the following hyperparameters: edge learning rate $\in \{0.03, 0.4, 0.8\}$, layer learning rate $\in \{0.001, 0.4, 0.8\}$, regularization edge learning rate $\in \{0.03, 0.4, 0.8\}$, regularization layer learning rate $\in \{0.001, 0.4, 0.8\}$, start edge sparsity $\in \{0.8, 0.9, 0.95\}$, target edge sparsity $\in \{0.975, 0.99, 1.05\}$, target layer sparsity $\in \{0.4, 0.69\}$, warmup steps $\in \{50, 250, 500\}$, disable node loss $\in \{\texttt{true}, \texttt{false}\}$, signs from EAP-IG $\in \{\texttt{true}, \texttt{false}\}$.

\begin{table*}[t]
\centering
\small
\setlength{\tabcolsep}{6pt}
\renewcommand{\arraystretch}{1.15}
\begin{tabularx}{\textwidth}{>{\raggedright\arraybackslash}X c c c}
\toprule
\makecell[l]{Parameter} &
\makecell[c]{GPT-2\\IOI} &
\makecell[c]{Qwen-2.5\\IOI} &
\makecell[c]{Qwen-2.5\\MCQA} \\
\midrule
Train steps            & 1000 & 1000 & 1000 \\
Batch                  & 20   & 10   & 10   \\
Warmup type            & \texttt{linear} & \texttt{linear} & \texttt{linear} \\
Warmup steps           & 250  & 250  & 250  \\
Disable node loss      & \(\checkmark\) & \(\times\) & \(\checkmark\) \\
Edge LR                & 0.4  & 0.8  & 0.4  \\
Layer LR               & 0.03 & 0.8  & 0.4  \\
Reg edge LR            & 0.8  & 0.8  & 0.4  \\
Reg layer LR           & 0.001 & 0.4 & 0.8 \\
Start edge spars.      & 0.90 & 0.95 & 0.95 \\
Target edge spars.     & 0.975 & 0.975 & 1.05 \\
Start layer spars.     & 0.0  & 0.0  & 0.0  \\
Target layer spars.    & 0.69 & 0.69 & 0.69 \\
Attribution / Signs    & z-score attribution & signs from EAP-IG & signs from EAP-IG \\
\bottomrule
\end{tabularx}
\caption{Final hyperparameters for the three runs. Columns are model--task combinations; rows are parameters. A checkmark (\(\checkmark\)) indicates a flag is enabled; a cross (\(\times\)) indicates disabled.}
\label{tab:final-hparams}
\end{table*}

\section{Experiments with IFR}
We also experimented with information flow routes, or IFR \cite{Ferrando2024a}, as a base method. The results were clearly weaker than the EAP-IG-inputs baseline, so we did not include this in our final submissions.

We need a special approach to compare edge scores from IFR with other base methods. This is because IFR is designed such that the scores of incoming edges to a given node sum to 1.
We therefore apply the same normalization to the scores from other methods as well, and proceed from there.
This also means that, when computing actual circuits from the final scores, we need greedy search, as already described by \citet{mueller2025mib}.

\end{document}